\documentclass[letterpaper]{article} 
\usepackage{aaai2026}  
\usepackage{times}  
\usepackage{helvet}  
\usepackage{courier}  
\usepackage[hyphens]{url}  
\usepackage{graphicx} 
\usepackage{natbib}  
\usepackage{caption} 
\usepackage{algorithm}
\usepackage{algorithmic}

\usepackage{subcaption} %
\usepackage{graphicx}
\usepackage{svg}
\usepackage{natbib}

\usepackage{mathrsfs}

\usepackage{amsmath}
\usepackage{amssymb}
\usepackage{amsthm}
\usepackage{mathtools}
\usepackage[mathscr]{eucal}%
\usepackage{bm}
\usepackage{relsize}
\usepackage{graphics}
\usepackage{tcolorbox}
\usepackage{multirow}
\usepackage{enumitem}
\usepackage{tablefootnote}
\usepackage{makecell}

\usepackage{array}
\usepackage{color, colortbl}
\definecolor{Gray}{gray}{0.9}
\definecolor{darkpastelgreen}{rgb}{0.01, 0.75, 0.24}
\definecolor{cadetgrey}{rgb}{0.57, 0.64, 0.69}
\definecolor{camel}{rgb}{0.76, 0.6, 0.42}
\definecolor{lightskyblue}{rgb}{0.53, 0.81, 0.98}
\definecolor{lightsblue}{rgb}{0.53, 0.81, 0.98}
\definecolor{lightblue}{rgb}{0.68, 0.85, 0.9}
\definecolor{softblue}{rgb}{0.85, 0.91, 0.98}
\definecolor{mygray}{gray}{0.6}
\usepackage{algorithm}
\usepackage{algorithmic}
\usepackage{pgf}
\usepackage{xinttools}
\usepackage{tikz}
\usetikzlibrary{arrows.meta}
\usepackage{textcomp}
\usetikzlibrary{calc,shapes,arrows,positioning,automata,trees}
\usepackage{placeins} 
\usepackage{tabularx}
\usepackage{graphicx}
\usepackage{listings}

\usepackage[font=small,labelfont=bf,tableposition=top]{caption}

\DeclareCaptionLabelFormat{andtable}{#1~#2  \&  \tablename~\thetable}

\usepackage{blindtext}

\usepackage{bigdelim}
\usepackage{colortbl} 
\definecolor{mColor1}{rgb}{0.95,0.95,0.95}

\usepackage{microtype}
\usepackage{graphicx}
\usepackage{booktabs} 
\usepackage{soul}

\definecolor{mydarkblue}{rgb}{0.68, 0.85, 1.0}
\definecolor{mydarkblue2}{rgb}{0,0.08,0.45}
\definecolor{mydarkblue3}{RGB}{151,204,255}
\definecolor{cvprblue}{rgb}{0.21,0.49,0.74}

\usepackage{newfloat}
\usepackage{listings}
\DeclareCaptionStyle{ruled}{labelfont=normalfont,labelsep=colon,strut=off} 
\floatstyle{ruled}
\newfloat{listing}{tb}{lst}{}
\floatname{listing}{Listing}
\title{VisChainBench: A Benchmark for Multi-Turn, Multi-Image Visual Reasoning Beyond Language Priors}
\author {
    Wenbo Lyu\textsuperscript{\rm 1},
    Yingjun Du\textsuperscript{\rm 2},
    Jinglin Zhao\textsuperscript{\rm 3},
    Xianton Zhen\textsuperscript{\rm 4},
    Ling Shao\textsuperscript{\rm 1}
}
\affiliations {
    \textsuperscript{\rm 1}University of Chinese Academy of Sciences\\
    \textsuperscript{\rm 2}University of Amsterdam\\
    \textsuperscript{\rm 2}Huazhong University of Science and Technology\\
    \textsuperscript{\rm 2}United Imaging Healthcare\\
    lvwenbo23@mails.ucas.ac.cn 
}
\usepackage{bibentry}

\begin{document}

\maketitle

\begin{abstract}
Understanding multi-image, multi-turn scenarios is a critical yet underexplored capability for Large Vision-Language Models (LVLMs). Existing benchmarks predominantly focus on static or horizontal comparisons—e.g., spotting visual differences or assessing appropriateness-while relying heavily on language cues. Such settings overlook progressive, context-dependent reasoning and the challenge of visual-to-visual inference.
To bridge this gap, we present VisChainBench, a large-scale benchmark designed to rigorously evaluate LVLMs’ ability to perform multi-step visual reasoning across sequential, interdependent tasks with minimal language guidance. VisChainBench contains 1,457 tasks spanning over 20,000 images across three diverse domains (e.g., daily scenarios, engineering troubleshooting), structured to mimic real-world decision-making processes. Uniquely, the benchmark is constructed using a multi-agent generation pipeline, ensuring high visual diversity and controlled language bias.
All the benchmark data and code for benchmark construction are available for viewing and download via following Link:
\item https://huggingface.co/datasets/eyehole/VisChainBench

\end{abstract}
\section{Introduction}

The rapid development of Large Vision-Language Models (LVLMs)~\cite{li2025benchmark,bai2025qwen2,blogIntroducingGemini,llavavlLLaVANeXTImproved,openaiHelloGPT4o} has led to impressive progress in multimodal understanding, enabling applications ranging from visual question answering~\cite{yu2023mm,masry2022chartqa} to interactive robotics~\cite{zitkovich2023rt,li2024manipllm}. A core capability underlying these advances is the ability to reason over multiple images across extended interactions-a skill essential for real-world tasks such as technical troubleshooting, intelligent device agents, or sequential visual decision-making. For instance, diagnosing a malfunctioning device may require analyzing its defective component, selecting the appropriate repair tool, and verifying the outcome—all through a chain of visual inputs.

In realistic scenarios such as embodied agents~\cite{yang2025embodiedbench,koh2024visualwebarena,jansen2024discoveryworld}, models are expected to track dynamic environmental changes across image sequences and respond accordingly. Similarly, personal assistant agents must interpret evolving screenshots and autonomously identify pending tasks. These applications require models to infer task objectives purely from visual context, without explicit textual prompts—a capability largely absent from current benchmark evaluations. This gap raises a critical question: what forms of reasoning must LVLMs perform to succeed in such settings? From assessing the current visual state to anticipating future steps and making decisions under uncertainty with less language guidance, robust multimodal reasoning is key.

\begin{figure}
    \centering
    \includegraphics[width=\linewidth]{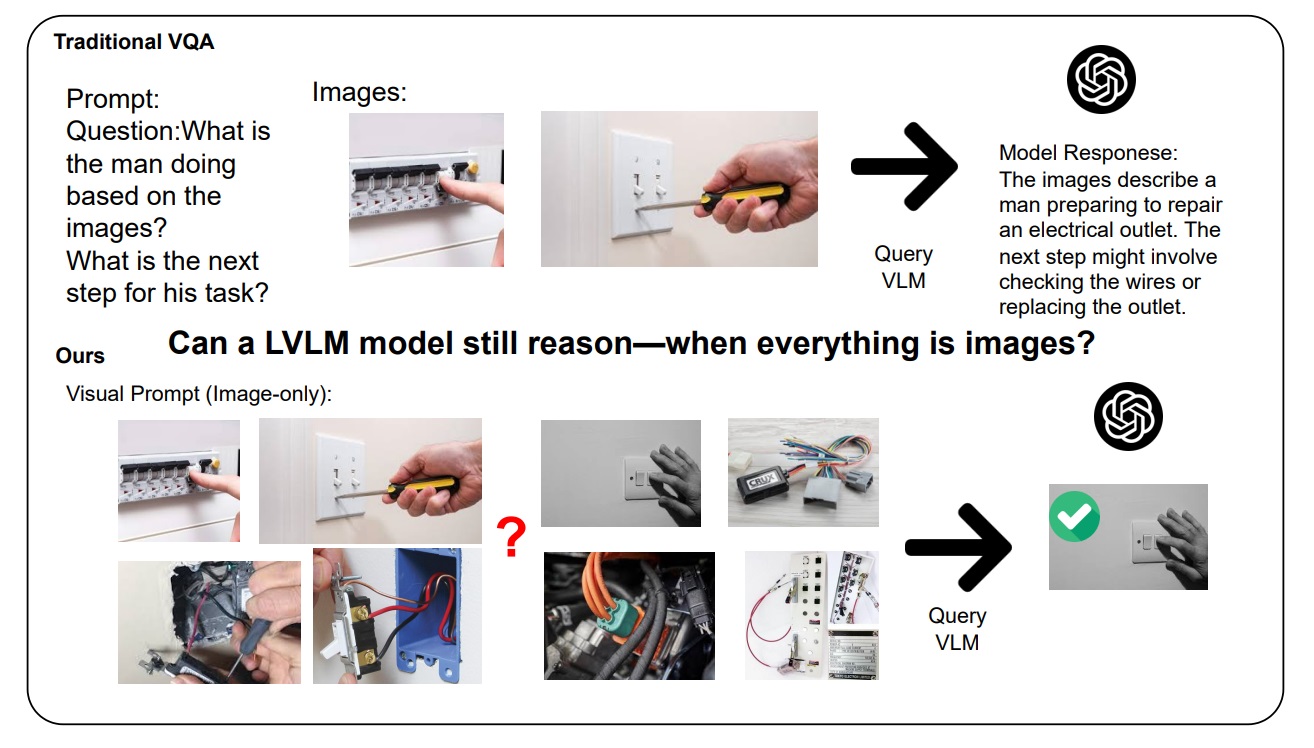}
    \caption{Traditional VQA tasks rely on language prompts and responses (top). VisChainBench introduces a purely visual paradigm, where the context, prompt, and answer are all images-challenging models to perform multi-step visual reasoning without textual cues (bottom).}
    \label{fig:enter-label}
\end{figure}


\begin{figure*}[t]
    \centering
    \includegraphics[width=\textwidth]{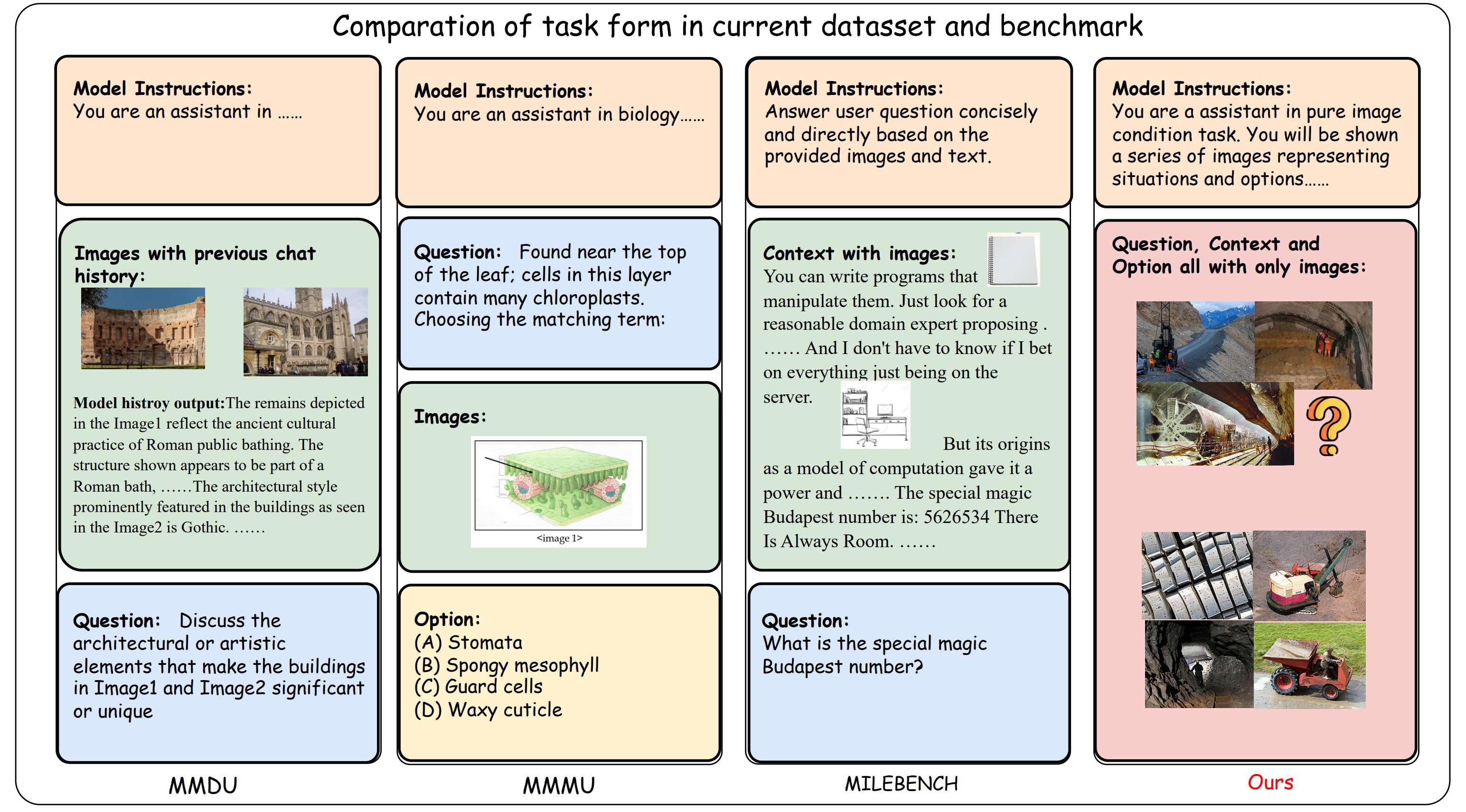}
    \caption{\textbf{Comparison of task formats in prior benchmarks.} Previous benchmarks are often text-heavy and encourage shallow image comparisons, whereas our benchmark emphasizes progressive image-grounded reasoning.}
    \label{fig:begin}
\end{figure*}


Although many LVLMs achieve strong results on perceptual or single-step tasks, their ability to conduct multi-step, image-to-image, and context-dependent reasoning, as mentioned above, remains underexplored. To address this, we introduce \textit{VisChainBench}, a benchmark specifically designed to evaluate multi-turn visual reasoning under minimal linguistic guidance. Unlike prior datasets, our benchmark presents tasks as structured visual chains, requiring models to reason progressively across linked visual contexts.

\textit{VisChainBench} is constructed using a multi-agent pipeline: initial task structures are generated by language models, relevant images are retrieved or synthesized, and human annotators refine both the tasks and their annotations. This process ensures high-quality, diverse, and visually grounded task sequences that challenge model reasoning beyond superficial patterns.

Our benchmark features the following distinctive aspects:  
\emph{(1) Multi-turn, multi-image structure.} Each task simulates extended human-AI interactions, involving up to 6 dialogue turns and 27 evolving images—capturing real-world complexity such as procedural workflows and troubleshooting steps.  
\emph{(2) Image-centric reasoning with distractors.} Tasks are designed to probe models' ability to reason between images, not merely identify isolated features, and include carefully controlled distractor options.  
\emph{(3) Long-context visual understanding.} With up to 27 images per task and over 20,000 images in total, the benchmark challenges models to process high-volume visual data with minimal textual cues.  
\emph{(4) Structured, reproducible evaluation.} All tasks are formatted as single- or multiple-choice questions with predefined ground-truth labels, and we release evaluation scripts to ensure consistency and comparability.

We evaluate eight proprietary and two open-source LVLMs on \textit{VisChainBench}. Our results reveal a substantial performance gap between proprietary and open models. To foster further progress, we release our benchmark construction pipeline to support the community in generating domain-specific data for training and evaluation.

\noindent\textbf{Our contributions are:}
\begin{itemize}
  \item We introduce a large-scale benchmark targeting multi-turn, multi-image visual reasoning under minimal language conditions. 
  \item We provide a systematic evaluation of existing LVLMs across diverse domains and reasoning formats.  
  \item We open-source our multi-agent dataset generation framework to promote future research in visual reasoning.
\end{itemize}

\section{Related Work}
\begin{figure*}[t]
    \centering
    \includegraphics[width=0.6\textwidth]{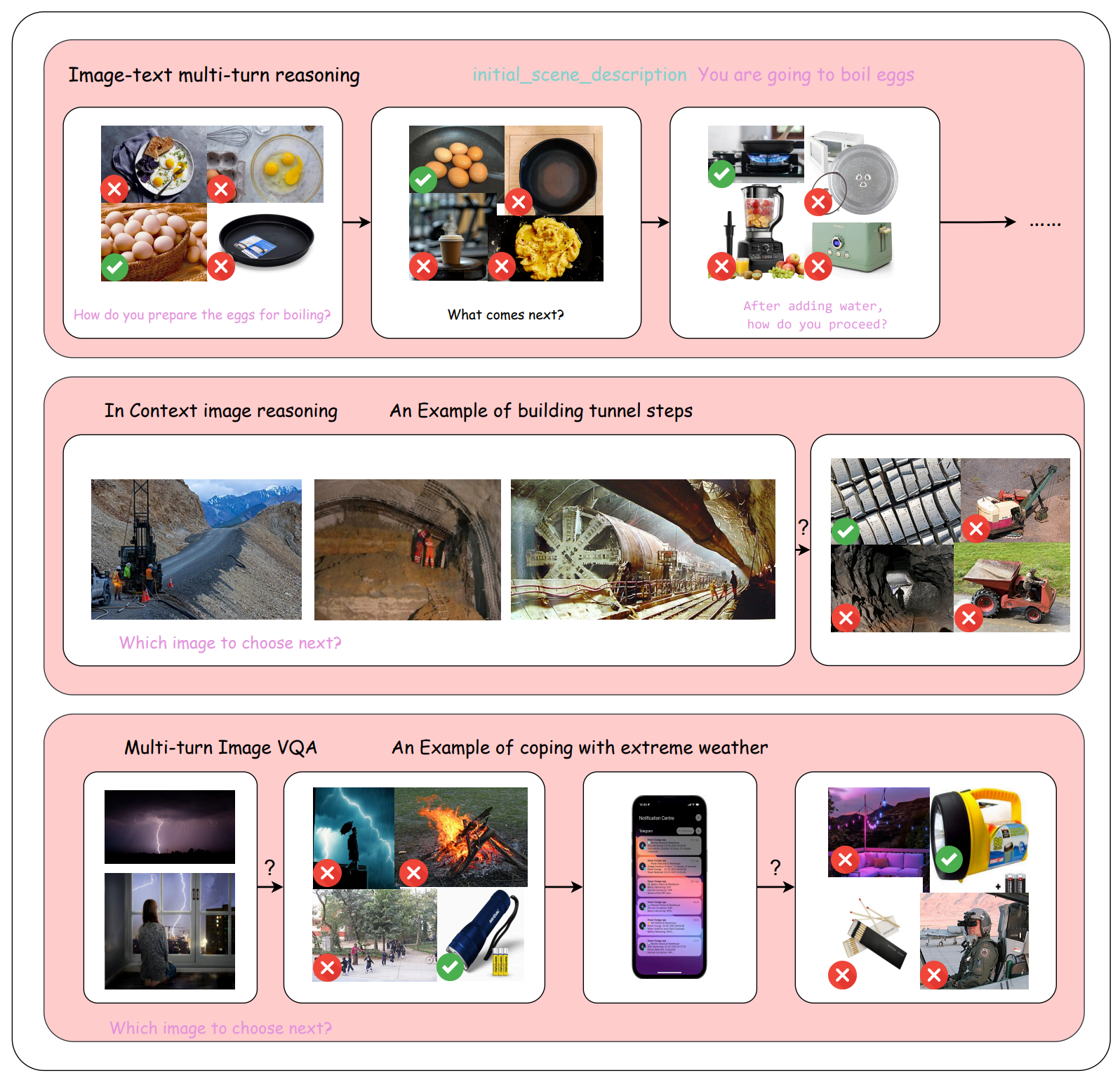}
    \caption{\textbf{Three kinds of reasoning example}, including Multi-turn Image-Text VQA, Image In-Context Reasoning and Multi-turn Image VQA. We keep the text instructions and questions in the Multi-turn Image-Text VQA but make the next-turn reasoning rely more on previous image choices. In Image In-Context Reasoning and Multi-turn Image VQA the tested models have to find out the current task instruction and question itself with only the task-agnostic prompt.}
    \label{fig:Example}
\end{figure*}
\paragraph{Multi-image and Long-Context MLLMs.}
The rapid advancements in Large Vision-Language Models(LVLMs) \cite{blogIntroducingGemini,abouelenin2025phi} have shown the ability to handle multiple image inputs and image-text sequences \cite{liu2024mia,lu2024deepseek}. However, some of the models have relatively short context windows for multi-image inputs. There are recent works that aim to overcome this shortcoming \cite{xue2024longvila,ge2024v2pe}. Some other LVLMs have developed the ability to perform video understanding \cite{cheng2024videollama,li2024llava,chen2024internvl}, and are trained to process multiple long image sequences and long contexts. Recently, researchers have focused on training LVLMs to comprehend multiple images using interleaved image-text corpora \cite{dong2024internlm,dong2024internlm4k,jiang2024maven,ye2024mplug}. However they are still trained with image-text pairs with instructions. The ability for LVLMs to reason under fewer text instructions has not been evaluated.

\paragraph{Current evaluation of LVLMs.}
Most of the popular LVLM benchmarks are evaluating the models on singe-image \cite{wu2023q,marino2019ok,goyal2017making,singh2019towards}. Recently there have been Benchmarks on testing models' multi-image processing abilities.Such as multi-image conversation \cite{liu2024mmdu}, multi-image understanding \cite{meng2024mmiu,tan2024devbench}, multi-image reasoning \cite{kazemi2024remi,zhao2024benchmarking}, multi-image comparation \cite{wu2024towards,wang2024muirbench} and multimodel multi-turn instruction \cite{liu2024convbench}. However, none of the existing benchmarks evaluated the combination of multi-image reasoning with image instruction and image question, highlighting a need for this framework. 

\paragraph{Comparison to Video Understanding Benchmarks.}
 While video understanding tasks~\cite{cheng2024videollama,wang2022internvideo,xue2024longvila,wang2024internvideo2} aim to evaluate temporal modelling and frame-level motion understanding, our benchmark focuses on high-level procedural reasoning across discrete, semantically meaningful visual steps. Each image in VisChainBench represents a distinct stage in a task workflow, such as "selecting a tool" or "verifying completion," rather than being a frame in a continuous video stream. This distinction makes VisChainBench more suitable for evaluating decision-making and planning capabilities, which are critical for embodied agents and multi-step assistants. In contrast to end-to-end video models, our benchmark encourages models to reason over image-to-image transitions and maintain long-term consistency across multi-turn interactions.

\paragraph{Compare to current multiturn multimodel agent bench} Compared to prior benchmarks such as EmbodiedBench~\cite{yang2025embodiedbench} and VisualAgentBench~\cite{liu2024visualagentbench}, which evaluate multimodal agents in grounded, task-oriented scenarios often with heavy reliance on textual prompts, our benchmark VisChainBench uniquely emphasizes multi-turn, multi-image reasoning with minimal language guidance. Specifically, VisChainBench introduces a purely visual reasoning paradigm, where the context, query, and answer are all image-based, enabling the assessment of visual-to-visual reasoning without language priors. This is an unevaluated capability for real-world agents operating in environments where explicit textual instructions are unavailable. These aspects make VisChainBench a complementary and necessary addition to the current ecosystem of multimodal multi-turn benchmarks.

\section{VisChainBench}
\subsection{Benchmark Overview}
In real-world applications, LVLMs capable of processing continuous image streams without relying on text-based prompts unlock transformative potential across industries. However, the development of standardised benchmarks for such image-stream-centric reasoning remains significantly lagging. 
This gap limits the systematic validation of LVLMs' ability to autonomously derive insights from raw image sequences, hindering their deployment in mission-critical domains like precision agriculture, infrastructure inspection, or emergency response systems, where dynamic visual reasoning must operate without human-curated prompts or post-hoc explanations.

Although many existing benchmarks claim to target multi-image and multi-turn dialogue scenarios, their evaluation protocols are predominantly text-centric, focusing on long-form language-based interactions. In most cases, only the initial input includes images, while subsequent turns rely heavily on textual exchanges with little or no additional visual input. As a result, the actual reasoning being tested is grounded in the initial image context but proceeds largely through linguistic prompts and responses. Benchmarks such as MMDU~\cite{liu2024mmdu} and MILEBench~\cite{song2024milebench} illustrate this trend, where task instructions and questions remain primarily language-driven. Consequently, these benchmarks offer limited evaluation of true image-to-image or non-text-based visual reasoning capabilities.

\begin{table}
\centering

\begin{tabular}{@{} l r r @{}}
\toprule
\textbf{Category} & \textbf{Metric} & \textbf{Count} \\
\midrule
Global Dataset 
    & Total Tasks & 1,457 \\
    & Total Images & 20,431 \\
    & Avg./Task & 14 \\
\midrule
Multi-turn-IT-VQA
    & Images & 9,826 \\
    & Avg. Images & 15 \\
    & Words/Q & 11 \\
\midrule
Image-In-Context 
    & Images & 3,192 \\
    & Avg./Task & 7 \\
\midrule
Multi-turn Image VQA 
    & Images & 7,413 \\
    & Avg./Task & 20 \\
\bottomrule
\end{tabular}
\caption{Statistics of VisChainBench}
\label{tab:stats_chart}
\end{table}

\begin{figure}
\centering
\includegraphics[width=0.2\textwidth]{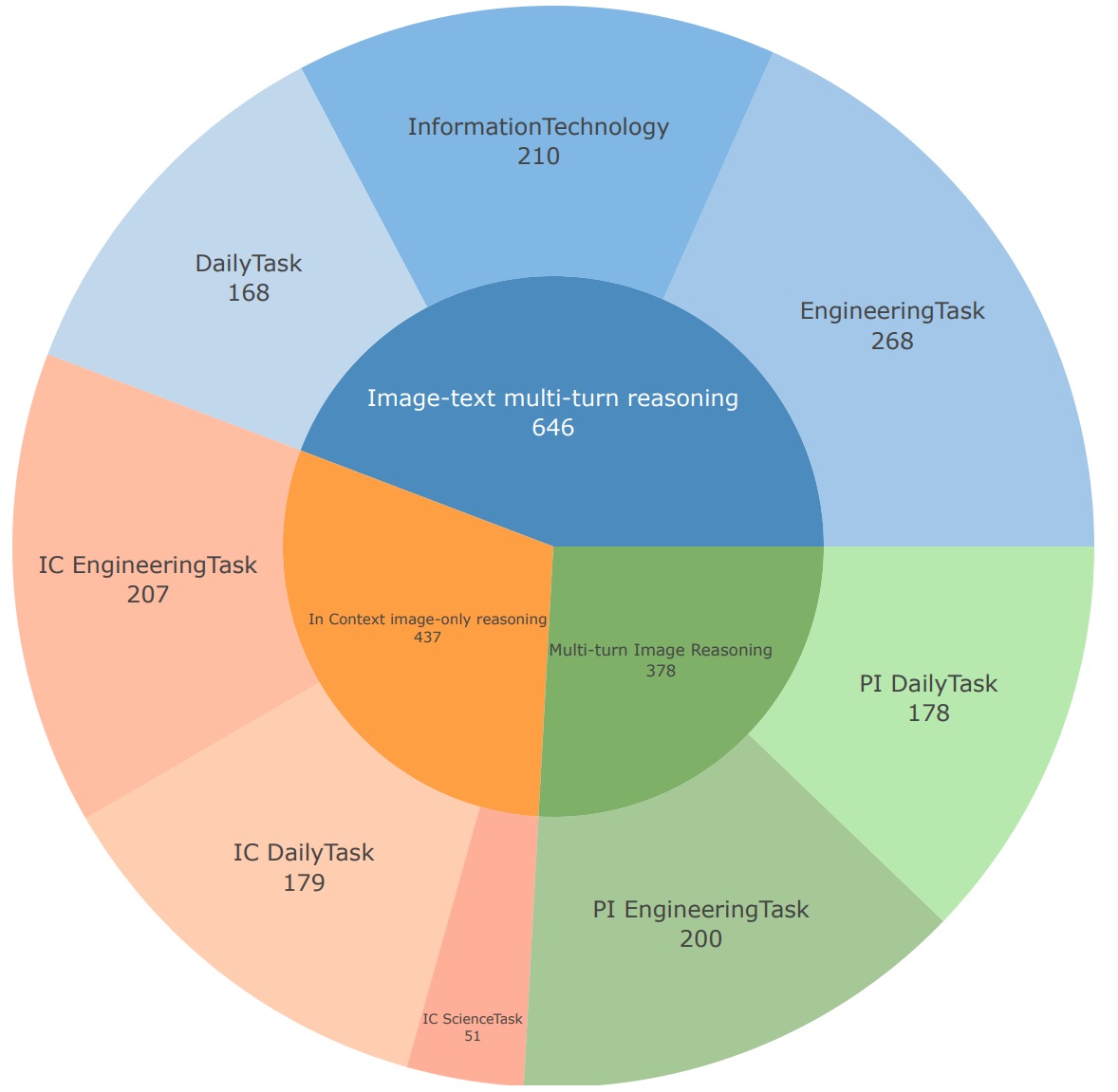}
\caption{\textbf{Task distribution across domains.} 
Our benchmark features tasks in three primary formats. 
Data was collected with emphasis on problem decomposition and solution strategies across diverse domains.}
\label{fig:pies}
\end{figure}

To evaluate the multi-image multi-turn lack of text-conditioned reasoning of current LVLMs, we developed a multi-agent framework and developed the VisChainBench based on the framework.
\subsection{Evaluation Taxonomy}
VisChainBench consists of three major evaluation forms: Image-text multi-turn reasoning, In Context image-only reasoning, and Image-only Multi-turn reasoning, as depicted in Figure \ref{fig:Example}. These tasks primarily construct the task instructions and questions using images as much as possible. The number of images contained in each subtask can be seen in Table \ref{tab:stats_chart}.

\noindent\textbf{Image-text multi-turn reasoning}
The Image-text multi-turn reasoning is designed to test the MLLM's ability to reason in multi-turn multi-model tasks, while the text question is designed to minimise the language hint and prior. The next turn question is designed to maximise the use of the former image-choosing trajectory, while minimising the beginning text instructions influence.
The Image-text multi-turn reasoning is constructed with four domains containing Daily task solving, Engineering task solving, Natural Science Understanding, and Information Technology Reasoning. Which totally contains 646 tasks, including 2k questions in multi-turn form.

\noindent\textbf{In Context image-only reasoning}
The In Context image-only reasoning is designed to test the MLLM's reasoning and context-understanding ability in the full image condition. The task is constructed without any text instruction, but using pure images as a condition brief and question formatting.
The In Context image-only reasoning is constructed with three domains containing Daily task understanding, Engineering task understanding, and Information Technology understanding, containing 437 tasks.

\noindent\textbf{Image-only Multi-turn reasoning}
The Image-only Multi-turn reasoning is designed to combine the In Context image-only reasoning into Image-text multi-turn reasoning, where we reformed the minimal text task description into image-only task description.This form makes the task harder and combines a mixture of image-to-image reasoning chains.
This task is constructed with three domains containing 437 tasks with over 7K images.

\subsection{Multi-Agent Benchmark Construction}
\begin{figure*}[t]
\includegraphics[width=0.7\textwidth]{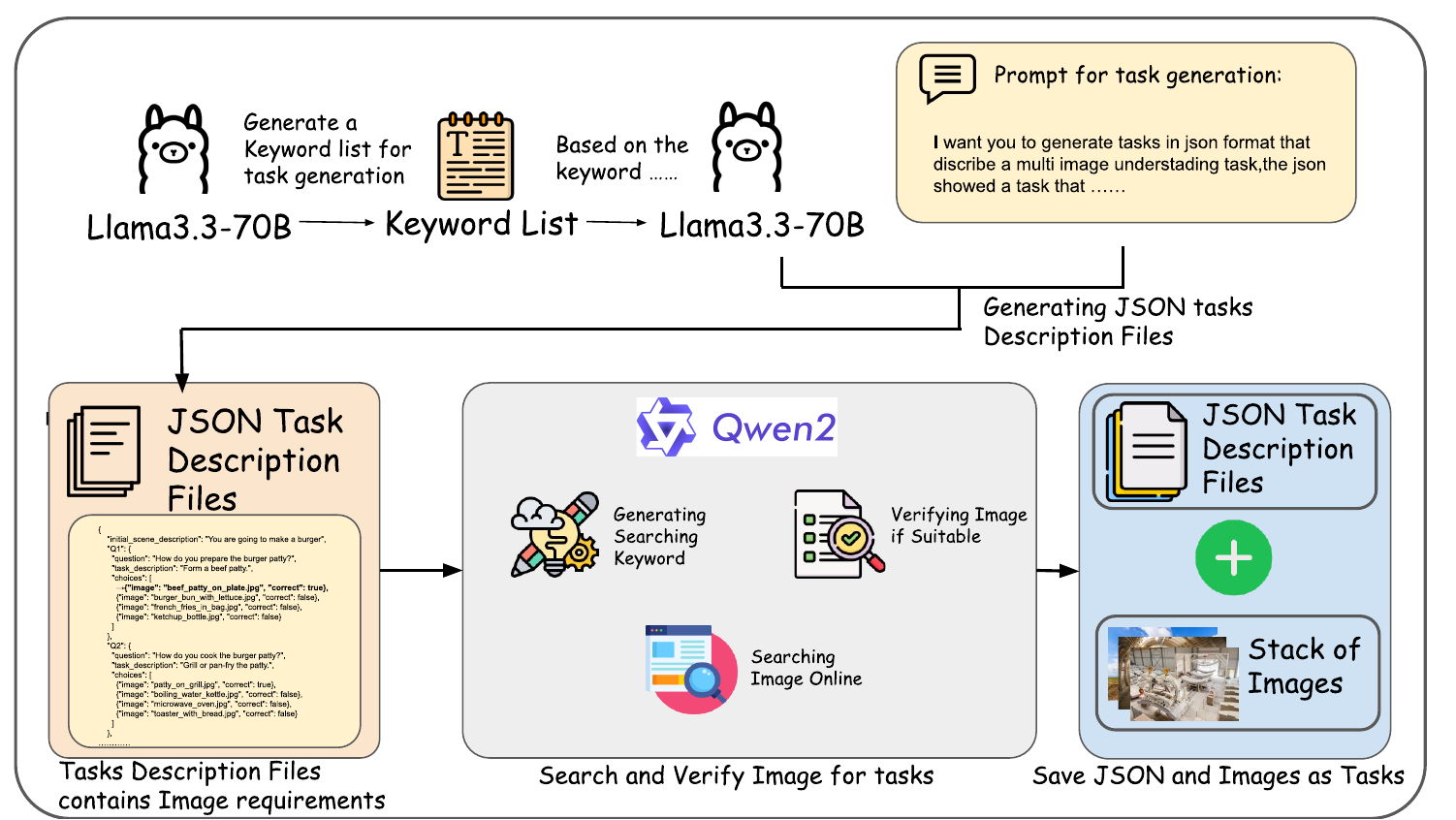}
\centering
\caption{\textbf{Benchmark Construction Process.}The workflow begins with automated task generation using \textbf{Llama3.3-70B} to produce structured JSON task descriptions. Then we use a \textbf{Qwen2-VL-72B} model to get corresponding images based on the Task description files and perform consistency verification. Validated tasks will be processed through human quality checks and modifications. This pipeline ensures systematic task creation while maintaining quality control between each processing stage.}
\centering
\end{figure*}
\noindent\textbf{Task generation}
The process begins with Llama3.3-70B \cite{llamaLlama} generating multi-step procedural tasks and decomposing them into sequential subtasks. For each subtask, the model constructs JSON-formatted task descriptions containing initial scene context,multi-turn visual questions,Contextually relevant answer choices with image slots, and Ground truth annotations for correct workflow progression.

\noindent\textbf{Image Collection}
To ensure visual consistency, we employ a dual-model verification system. A Qwen2-VL-72B model first generates search keywords from task descriptions. Initial images are retrieved via keyword-based searches from the internet. We filter sources for commercial licenses, retaining only images with Creative Commons licenses. A second Qwen2-VL-72B then performs contextual validation, rejecting mismatched images. For rejected cases, we conduct multiple rounds of refined keyword searches before considering synthetic image generation. Only if suitable images remain unavailable after these repeated search efforts do we utilize doubao-t2i-drawing \cite{bytedanceSeedreamTechnical} to generate synthetic images via Qwen model-driven prompt engineering.Analysis of 500 sampled images from our task corpus shows that less than 5\% are T2I-generated. This iterative validation loop continues until all images match the task context.

\noindent\textbf{Task vertification} The synthetic dataset undergoes automated validation using Qwen2-VL-72B to assess task integrity. The model processes the entire dataset through multi-turn reasoning, generating predicted answers for each visual question. These predictions are compared against the ground truth labels. The system aggregates raw correctness metrics and identifies discrepancies between model outputs and human annotations. These results are compiled into an error priority list that highlights tasks requiring human review, focusing on cases where model predictions conflict with ground truth labels. The verification outputs guide human annotators by flagging potential inconsistencies while preserving the original dataset's evaluation objectives.

\noindent\textbf{Human Annotators and Quality Control}
To ensure high-quality human annotations while minimising labour, we adopt a three-step verification and correction pipeline.\textbf{(1)} We conduct an initial round of human review to filter out low-quality or inconsistent tasks. Subsequently, annotators answer the filtered tasks and flag any remaining ambiguities or errors.\textbf{(2)} We develop a specialised user interface (UI) to visualise correctness metrics, integrating raw performance data from Large Vision-Language Models (LVLMs) with human annotations. This UI enables systematic comparison, helping human experts efficiently review and refine the dataset.\textbf{(3)} To validate annotation reliability, a separate group of annotators completes a quiz-based evaluation of the revised dataset. If task accuracy falls below a predefined threshold, the dataset is reverted to the second step for further corrections. This iterative process ensures robust quality control while balancing annotation effort and precision.

During human annotation, all our annotators were Master-level or junior Phd-level students, with a total of 6 participants. They all have relevant with professional backgrounds. All of the annotators are trained with a small subset of pre-annotated good dataset examples.
\subsection{Evaulation}
Our dataset uses a hard-coded answer label in the JSON file. All the LVLMs are tested with the same prompt. We have tested different image input methods with different label methods tries, in order to give the LVLMs the instructions of the image order. The LVLMs are instructed to answer the image label.
\section{Experiments}
\label{sec: experiment}
We evaluated some representative LVLMs on our VisChainBench benchmark and presented the analysis of our findings. 
\subsection{Experimental setup}\label{sec:implementation}

We evaluated the performance of two closed-source API models, GPT-4o \cite{openaiHelloGPT4o} and gemini-2.0-flash \cite{blogIntroducingGemini}.For open-source MLLMs we choose Qwen2.5VL 32B/7B/3B \cite{bai2025qwen2},Phi4-multimodel-instruct \cite{abouelenin2025phi},MiniCPM-V 2.6 \cite{minicpm},LLAVA-NEXT-34b \cite{llavavlLLaVANeXTImproved},llama3.2-vision:11b-instruct \cite{metaLlama32},InternVL3-14B/7B\cite{zhu2025internvl3}. 

We all use the same prompt structure during testing the MLLMs, with a zero-shot setting, the default text prompt is detailed in the appendix. The image for choosing will be shown with a corner number label, indicating the choice sequence, following the findings in \cite{li2024monkey}. We also tested a long thinking model, VLM-R1 \cite{shen2025vlmR1}, the results show there is no significant improvement for this task.
\subsection{Experimental Result and Analysis}
\begin{table*}

  \label{sample-table}
  \centering
  \begin{tabular}{llcccccc}
    \toprule
    \multicolumn{2}{c}{} & \multicolumn{5}{c}{Part} \\
    \cmidrule(lr){3-7}
    Models & Param & \multicolumn{2}{c}{ITMR} & ICIR & \multicolumn{2}{c}{IOMR} & Overall \\
    \cmidrule(lr){3-4} \cmidrule(lr){5-5} \cmidrule(lr){6-7}
     &  & CA & TC & TC & CA & TC & \\
    \midrule
    \multicolumn{8}{l}{\textit{Closed-source models:}} \\
    gpt-4o &  & 77.65& 31.58& \textbf{71.74}& 75.75 &  \textbf{30.01 }& \textbf{73.85}\\ 
    gemini-2.0-flash &  &\textbf{ 82.04}& \textbf{46.10}& 70.73& \textbf{75.81}& 19.24& 67.95\\ 
    \hline
    \multicolumn{8}{l}{\textit{Open-source models:}} \\
    LLAVA-NEXT:34b & 34B & 10.85 & 0 & 10.28 & 25.32& 0& 19.72\\ 
    Qwen2.5VL32B & 32B & \textbf{71.42} & \textbf{29.93} & 25.91 &\textbf{57.93} & \textbf{12.04}& 51.97\\
    InternVL3-14B & 14B & 65.67 & 23.0 & \textbf{33.23} & 57.74& 9.99& \textbf{52.21}\\ 
    llama3.2-vision:11B-it & 11B & 7.25 & 0.25 & 10.79 & 28.67& 1.06& 15.57\\
    InternVL3-7B & 7B & 56.60 & 6.63 & 24.04 & 46.65& 5.15& 42.43\\ 
    MiniCPM-V 2.6  & 8B  & 23.57 & 2.40 & 17.50 & 46.69& 4.50& 25.01\\ 
    Phi4-multimodel-it  & 6B  & 25.48 & 1.36 & 10.57 & 17.78& 0.53& 17.94\\ 
    Qwen2.5VL7B  & 7B  & 54.44& 8.86 & 20.88 & 33.91& 1.94& 35.56\\ 
    Qwen2.5-VL-3B-it& 3B  & 30.45& 1.14& 2.65& 33.81& 1.31& 22.30\\ 
    Qwen2.5VL-3B-VLM-R1 & 3B  & 26.35& 2.36& 3.18& 37.33& 1.13& 22.29\\ 
    \bottomrule
  \end{tabular}
     \caption{\textbf{Experiment Result on VisChainBench}.We report the metrics of Correct Answered Questions percentage (CA) and Task Completed percentage (TC) in three task forms: Image-Text Multi-turn Reasoning (ITMR), In Context Image-only Reasoning (ICIR) and Image-Only Multi-turn Reasoning (IOMR). For Overall, we simply average the CA score in three tasks. The CA and TC are the same since ICIR only involves one round of questioning, which we use TC to calculate the overall CA.}
\end{table*}
Closed-source models (GPT-4o, Gemini) outperform open-source models by a large margin, especially in ITMR and ICIR.Gemini-2.0-flash achieves the highest CA (82.04\%) in ITMR, while GPT-4o leads in ICIR (71.74\%). We found for MLLMs with smaller parameters, the performance on multi-turn multi-image reasoning shows a huge gap compare to the models with Bigger parameters. Larger models (e.g., Qwen2.5VL-32B) significantly outperform smaller variants (e.g., Qwen2.5VL-3B), with a 41\% CA gap in ITMR.

For open source models, the strong performance of models like Qwen2.5VL-32B and InternVL3-14B on VisChainBench can be attributed to their training on structured, multi-image, and instruction-aligned data, as well as architectural designs that support long-context visual reasoning. These models are better equipped for image-to-image inference and sequential visual understanding, which are central to our benchmark. In contrast, models such as LLaVA-NEXT, MiniCPM, and Phi4 underperform due to their focus on single-image tasks, limited context handling, and reliance on language-heavy instruction tuning. Their pretraining lacks sufficient exposure to procedural or multi-turn visual workflows, making them less capable in image-only or low-text reasoning settings. This highlights the importance of targeted pretraining for multi-step visual reasoning under minimal language guidance.

\begin{figure}
    \centering
    \includegraphics[width=0.4\textwidth]{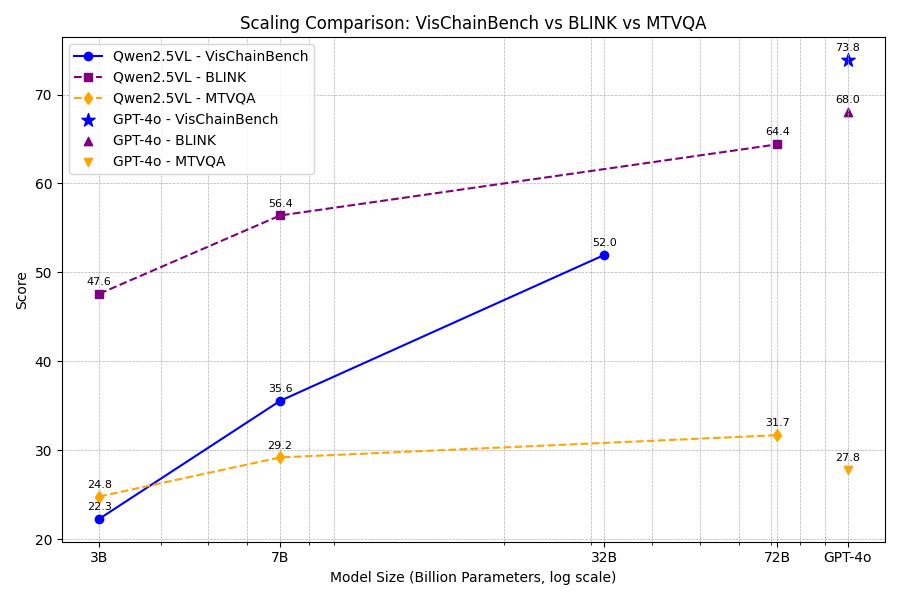}
    \caption{Scaling Comparison of Qwen Models on VisChainBench, BLINK and MTVQA Tasks}
    \label{fig:model_comparison}
\end{figure}

We observe a steep performance increase in Fig.\ref{fig:model_comparison} for vision-language models as model size increases—much steeper than scaling trends reported on classical benchmarks\cite{bai2025qwen2,zhu2025internvl3} like BLINK\cite{fu2024blink} and MTVQA\cite{tang2024mtvqa}. For instance, Qwen2.5VL improves by over 30 points from 3B to 32B on VisChainBench, while comparable gains on VQA tasks typically require jumps from 7B to 70B (e.g., BLIP-2 to Gemini). GPT4o also follows this trend.

We hypothesize this is due to the nature of our benchmark: multi-turn, multi-image reasoning tasks demand latent world model construction and temporal-spatial coherence, capabilities which emerge only at larger scales. This is consistent with scaling trends observed in math reasoning (e.g., GSM8K) and video understanding (e.g., NExT-QA), where task structure—not dataset scale—is the primary bottleneck at small model sizes.

The long thinking model (VLM-R1) with a small parameter size shows no observable advantage on VisChainBench. In fact, the specialised VLM-R1 variant of Qwen2.5VL-3B performs slightly worse than the base model across all sub-tasks. This suggests that long-thinking fine-tuning, at least in its current form, may not enhance the image-to-image reasoning capabilities required for multi-turn visual tasks. Upon inspecting the model outputs, we observed that VLM-R1 often bypasses the expected multi-step reasoning process and directly outputs an answer without generating explicit intermediate reasoning chains. This behavior indicates that, despite being optimized for chain-of-thought prompting, the model may not have learned to apply such reasoning patterns in visual-only or low-text scenarios. These findings raise important questions about the generalization of long-thinking alignment methods from language-dominant tasks to vision-centric benchmarks.

We evaluate the impact of Chain-of-Thought (CoT) prompting~\cite{wei2022chain} on Qwen2.5VL-32B’s performance on the Daily Task subset (Table~\ref{ablation-table}), focusing on three sub-tasks: Image-Text Multi-turn Reasoning (ITMR-Daily), In-Context Image-only Reasoning (ICIR-Daily), and Image-only Multi-turn Reasoning (IOMR-Daily). The results show significant gains on tasks with text instructions: +6.95\% CA (66.58\% → 73.53\%) and +9.52\% TC (20.24\% → 29.76\%) in ITMR-Daily, demonstrating CoT’s effectiveness when linguistic scaffolding is present. However, for low-text or image-only tasks such as ICIR-Daily, the gains are marginal (22.91\% vs. 22.35\% TC), suggesting that CoT provides little additional value when visual context must be reasoned over without textual support. Interestingly, for IOMR-Daily, which combines image-only and multi-turn reasoning, CoT still brings some improvement, hinting that chain-style reasoning can help even in reduced-text scenarios if the model is already trained for multi-turn understanding. We hypothesise that CoT’s limited impact in image-only tasks may stem from the vision encoder's inability to explicitly "see" reasoning chains between visual tokens—unlike in text, where token-level reasoning can be more easily structured. This raises the open question of whether visual CoT requires fundamentally architectural support or fine-tuning objectives to be effective.

\begin{table}
  \label{ablation-table}
  \centering
  \begin{tabular}{lccccc}
    \toprule
    & \multicolumn{2}{c}{ITMR-Daily} & ICIR-Daily & \multicolumn{2}{c}{IOMR-Daily}\\
    \cmidrule(lr){2-3} \cmidrule(lr){4-4} \cmidrule(lr){5-6}
     & CA & TC & TC & CA & TC\\
    \midrule
    Zero-shot & 66.58 & 20.24 & \textbf{22.91} &64.55 & 18.54\\
    with CoT & \textbf{73.53} & \textbf{29.76} & 22.35 &\textbf{70.84} &\textbf{24.16}\\
    \bottomrule
  \end{tabular}
     \caption{\textbf{Ablation Study on Qwen2.5VL32B}. We analyse the performance of Chain-of-Thought prompting on Image-Text Multi-turn Reasoning (ITMR) and In Context Image-only Reasoning (ICIR) tasks on the Daily Task Reasoning subset, reporting Correct Answered Questions percentage (CA) and Task Completed percentage (TC).}
  \label{fig:cot}
\end{table}
\begin{figure}[t]
    \centering
    \begin{subfigure}[t]{0.95\columnwidth}
        \centering
        \includegraphics[width=0.32\columnwidth]{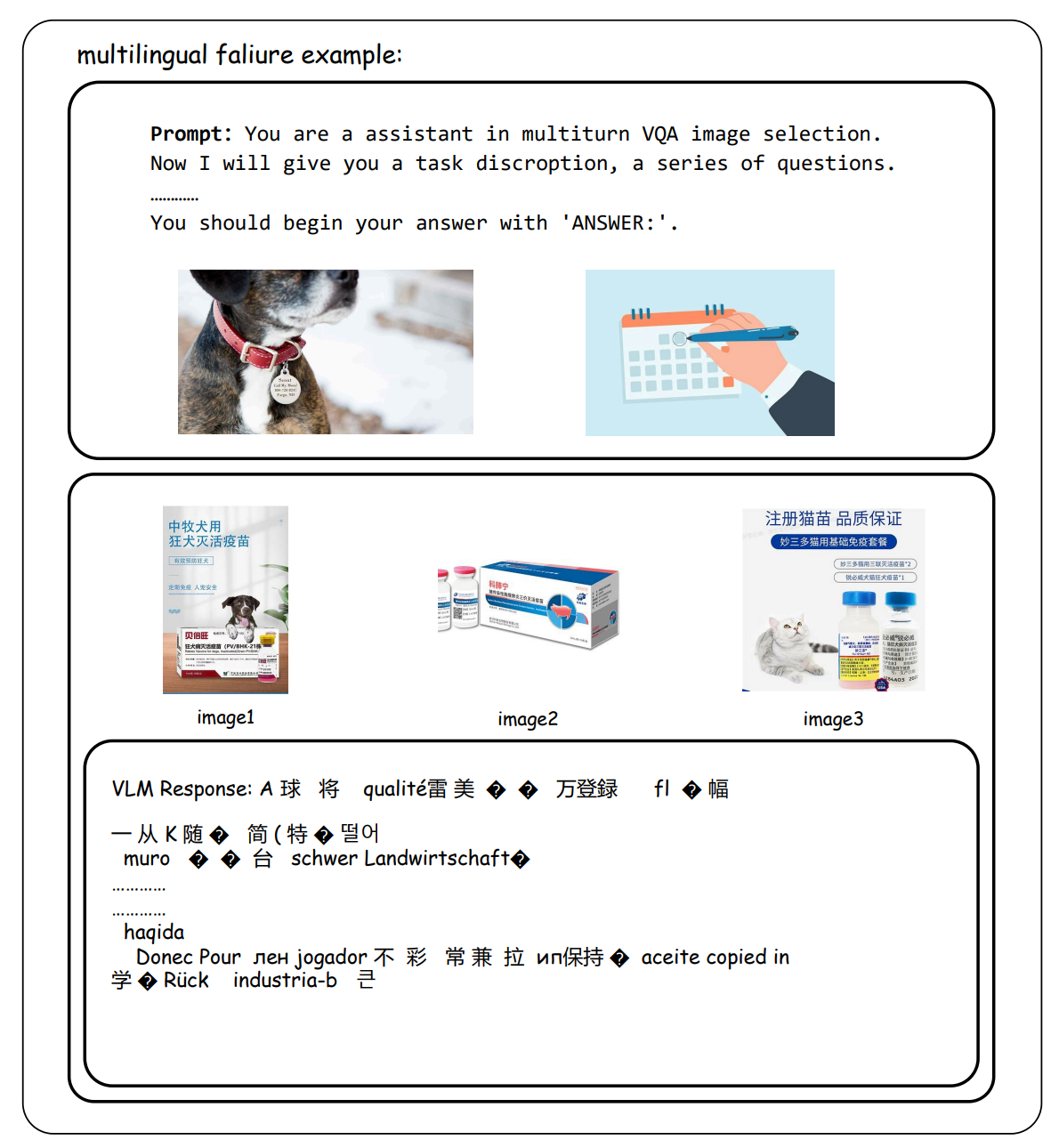}
        \includegraphics[width=0.32\columnwidth]{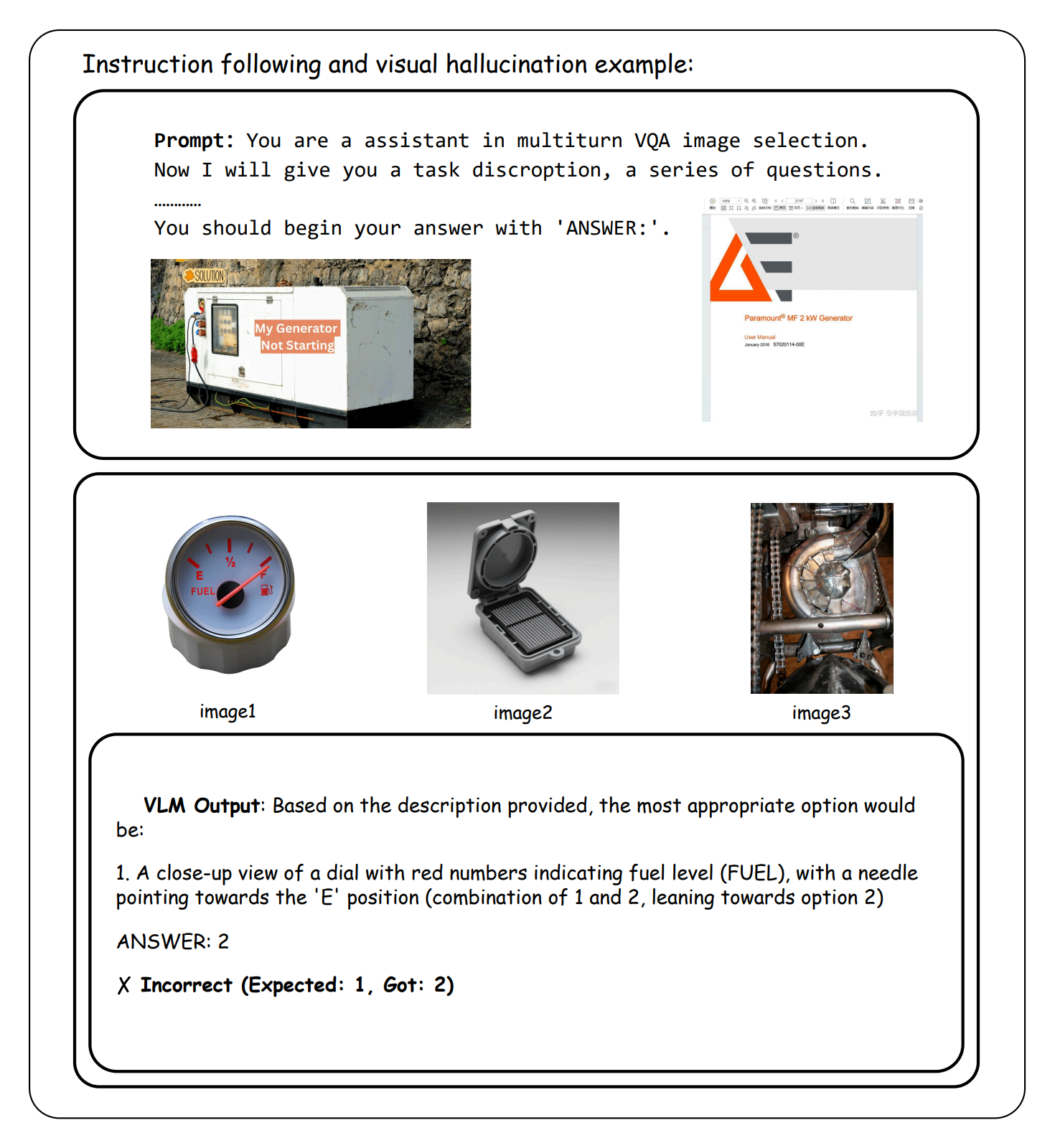}
        \includegraphics[width=0.32\columnwidth]{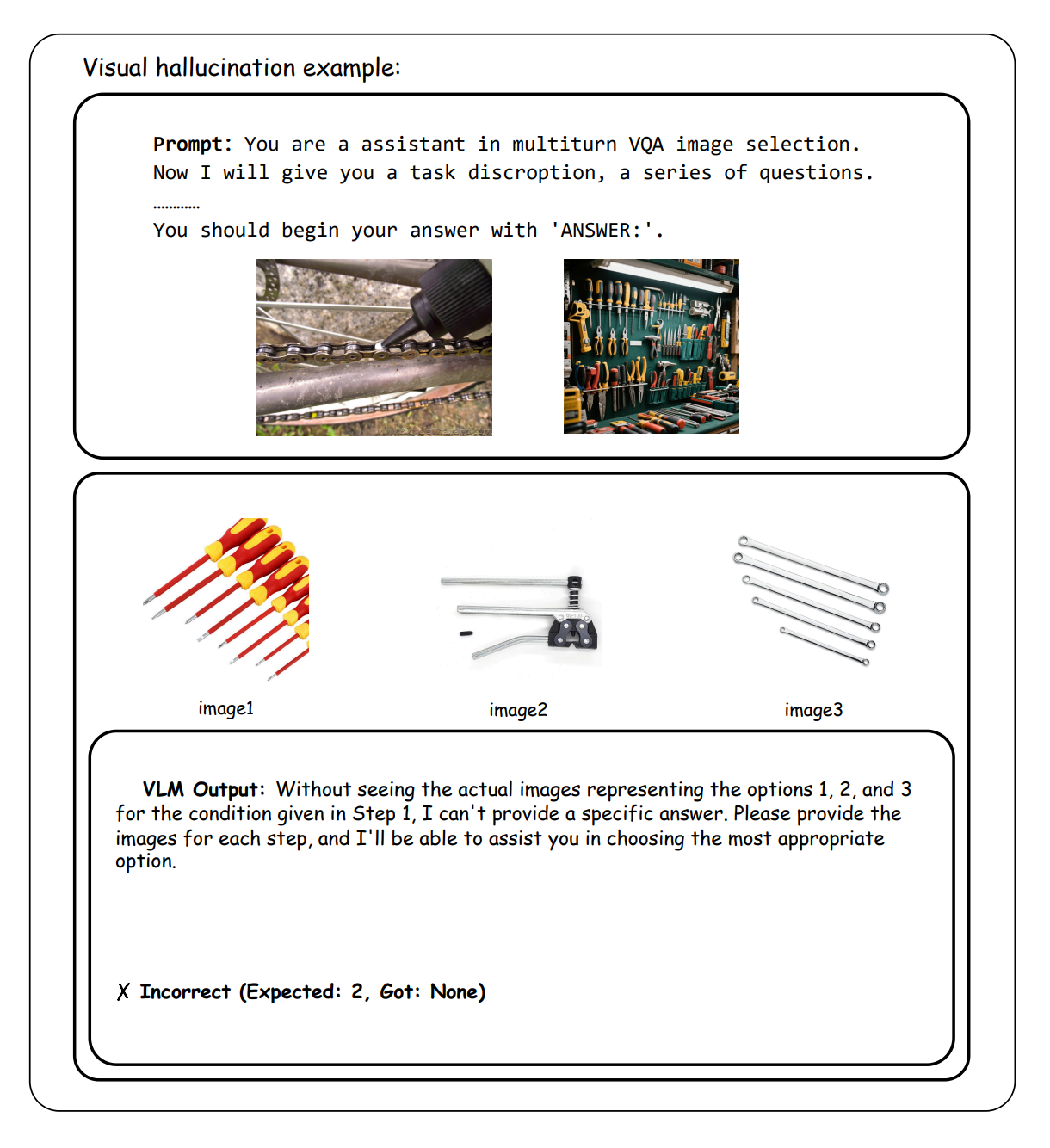}
        \caption*{}
    \end{subfigure}
    \caption{\textbf{Model Failure Cases.} Persistent limitations in LVLMs observed during multi-image reasoning. Left: Phi4 multilingual hallucination. Center: Qwen2.5VL-7B visual hallucination. Right: Phi4 visual hallucination. Readers are encouraged to zoom in for finer detail.}
    \label{fig:failures}
\end{figure}

\subsection{Error Case Study}
In evaluating model performance, we observe several critical issues that hinder effective assessment. Some models exhibit instruction-following deficiencies, such as LLaVA, which, despite correctly identifying the right answers, fail to adhere to the specified output formats. Additionally, certain models demonstrate refusal behaviours by requesting additional details before generating outputs, potentially stemming from over-reliance on input completeness. More critically, hallucination remains a pervasive problem, with models frequently misinterpreting visual content, such as misidentifying objects or their logical relationships, or missing image inputs in Fig.\ref{fig:failures}. Moreover, for low-performance models in our task, severe chaotic outputs and factual inaccuracies emerge, particularly in multilingual multimodal scenarios. For instance, Phi-4 exhibits erratic behaviours when processing cross-lingual inputs, such as conflating semantic meanings across languages or generating contradictory responses in reasoning steps. These systemic shortcomings collectively underscore significant challenges in current vision-language models, particularly in instruction compliance, multimodal recognition, and visual factual reasoning.

\section{Conclusion}
\label{sec: conclusion}

By leveraging a multi-agent framework for task generation, image collection, and validation, VisChainBench ensures visual complexity and contextual consistency across 1.4k tasks spanning three diverse domains. The benchmark’s design emphasises image-centric reasoning over text-based cues, closely mirroring real-world scenarios such as technical troubleshooting and sequential decision-making. Human annotation, guided by systematic quality control mechanisms, further ensures task integrity and relevance.

This work underscores the need for benchmarks that prioritise visual-to-visual reasoning in LVLM development, particularly for applications requiring dynamic interaction with visual environments. Future research directions include expanding the benchmark into more domains, for instance, medical procedures or mathematical reasoning could be a promising area for exploration. Another possible direction is exploring new interaction methods with current LVLMs. Besides the traditional text and image-based interactions, we can investigate more intuitive and natural ways of communication. For example, gesture-based interaction LVLMs or a self-interested ( through vision ) multimodal agent could be an interesting direction.

\noindent\textbf{Limitations}   We acknowledge three key limitations in our benchmark design. \textbf{(1)} VisChainBench primarily evaluates LVLMs in three specific domains (daily tasks, engineering scenarios, and information technology contexts), potentially overlooking specialised domains like medical procedures, mathematical reasoning, or artistic creation that demand distinct reasoning patterns. \textbf{(2)} While our benchmark is designed to minimise language usage, a limited amount of text is still employed in the task prompts to initialise the model’s behaviour and define output formats. This necessary scaffolding—such as the instruction to begin answers with 'ANSWER:' or to select from labelled options—may still introduce subtle language priors or instruction-following biases. As a result, it is not a fully language-free evaluation, and future work may explore fully instruction-less prompting or alternative interface paradigms.



\bibliography{ref}

@article{liu2024mmdu,
  title={Mmdu: A multi-turn multi-image dialog understanding benchmark and instruction-tuning dataset for lvlms},
  author={Liu, Ziyu and Chu, Tao and Zang, Yuhang and Wei, Xilin and Dong, Xiaoyi and Zhang, Pan and Liang, Zijian and Xiong, Yuanjun and Qiao, Yu and Lin, Dahua and Others},
  journal={arXiv preprint arXiv:2406.11833},
  year={2024}
}

@article{song2024milebench,
  title={Milebench: Benchmarking mllms in long context},
  author={Song, Dingjie and Chen, Shunian and Chen, Guiming Hardy and Yu, Fei and Wan, Xiang and Wang, Benyou},
  journal={arXiv preprint arXiv:2404.18532},
  year={2024}
}

@article{yang2025embodiedbench,
  title={Embodiedbench: Comprehensive benchmarking multi-modal large language models for vision-driven embodied agents},
  author={Yang, Rui and Chen, Hanyang and Zhang, Junyu and Zhao, Mark and Qian, Cheng and Wang, Kangrui and Wang, Qineng and Koripella, Teja Venkat and Movahedi, Marziyeh and Li, Manling and Others},
  journal={arXiv preprint arXiv:2502.09560},
  year={2025}
}

@article{bai2025qwen2,
  title={Qwen2. 5-vl technical report},
  author={Bai, Shuai and Chen, Keqin and Liu, Xuejing and Wang, Jialin and Ge, Wenbin and Song, Sibo and Dang, Kai and Wang, Peng and Wang, Shijie and Tang, Jun and Others},
  journal={arXiv preprint arXiv:2502.13923},
  year={2025}
}

@inproceedings{fu2024blink,
  title={Blink: Multimodal large language models can see but not perceive},
  author={Fu, Xingyu and Hu, Yushi and Li, Bangzheng and Feng, Yu and Wang, Haoyu and Lin, Xudong and Roth, Dan and Smith, Noah A and Ma, Wei-Chiu and Krishna, Ranjay},
  booktitle={European Conference on Computer Vision},
  pages={148--166},
  year={2024},
  organization={Springer}
}

@article{tang2024mtvqa,
  title={Mtvqa: Benchmarking multilingual text-centric visual question answering},
  author={Tang, Jingqun and Liu, Qi and Ye, Yongjie and Lu, Jinghui and Wei, Shu and Lin, Chunhui and Li, Wanqing and Mahmood, Mohamad Fitri Faiz Bin and Feng, Hao and Zhao, Zhen and Others},
  journal={arXiv preprint arXiv:2405.11985},
  year={2024}
}

@article{zhu2025internvl3,
  title={Internvl3: Exploring advanced training and test-time recipes for open-source multimodal models},
  author={Zhu, Jinguo and Wang, Weiyun and Chen, Zhe and Liu, Zhaoyang and Ye, Shenglong and Gu, Lixin and Tian, Hao and Duan, Yuchen and Su, Weijie and Shao, Jie and Others},
  journal={arXiv preprint arXiv:2504.10479},
  year={2025}
}

@article{liu2024visualagentbench,
  title={Visualagentbench: Towards large multimodal models as visual foundation agents},
  author={Liu, Xiao and Zhang, Tianjie and Gu, Yu and Iong, Iat Long and Xu, Yifan and Song, Xixuan and Zhang, Shudan and Lai, Hanyu and Liu, Xinyi and Zhao, Hanlin and Others},
  journal={arXiv preprint arXiv:2408.06327},
  year={2024}
}

@article{shen2025vlmR1,
  title={Vlm-r1: A stable and generalizable r1-style large vision-language model},
  author={Shen, Haozhan and Liu, Peng and Li, Jingcheng and Fang, Chunxin and Ma, Yibo and Liao, Jiajia and Shen, Qiaoli and Zhang, Zilun and Zhao, Kangjia and Zhang, Qianqian and Others},
  journal={arXiv preprint arXiv:2504.07615},
  year={2025}
}

@misc{openaiHelloGPT4o,
	author = {{openai}},
	title = {{H}ello {G}{P}{T}-4o --- openai.com},
	howpublished = {\url{https://openai.com/index/hello-gpt-4o/}},
	year = {{2024}},
	note = {[Accessed 29-04-2025]},
}

@misc{blogIntroducingGemini,
	author = {{Google}},
	title = {{I}ntroducing {G}emini 2.0: our new {A}{I} model for the agentic era --- blog.google},
	howpublished = {\url{https://blog.google/technology/google-deepmind/google-gemini-ai-update-december-2024/#ceo-message}},
	year = {{2024}},
	note = {[Accessed 29-04-2025]},
}

@article{abouelenin2025phi,
  title={Phi-4-mini technical report: Compact yet powerful multimodal language models via mixture-of-loras},
  author={Abouelenin, Abdelrahman and Ashfaq, Atabak and Atkinson, Adam and Awadalla, Hany and Bach, Nguyen and Bao, Jianmin and Benhaim, Alon and Cai, Martin and Chaudhary, Vishrav and Chen, Congcong and Others},
  journal={arXiv preprint arXiv:2503.01743},
  year={2025}
}

@misc{minicpm,
	author = {{OpenBMB}},
	title = {{Y}our connected workspace for wiki, docs \& projects | {N}otion --- openbmb.notion.site},
	howpublished = {\url{https://openbmb.notion.site/MiniCPM-o-2-6-A-GPT-4o-Level-MLLM-for-Vision-Speech-and-Multimodal-Live-Streaming-on-Your-Phone-185ede1b7a558042b5d5e45e6b237da9}},
	year = {{2024}},
	note = {[Accessed 29-04-2025]},
}

@misc{llavavlLLaVANeXTImproved,
  author = {Liu, Haotian and Li, Chunyuan and Li, Yuheng and Li, Bo and Zhang, Yuanhan and Shen, Sheng and Lee, Yong Jae},
  title = {LLaVA-NeXT: Improved reasoning, OCR, and world knowledge --- llava-vl.github.io},
  howpublished = {\url{https://llava-vl.github.io/blog/2024-01-30-llava-next/}},
  year = {2024},
  note = {[Accessed 29-04-2025]},
}

@misc{metaLlama32,
	author = {{meta}},
	title = {{L}lama 3.2: {R}evolutionizing edge {A}{I} and vision with open, customizable models --- ai.meta.com},
	howpublished = {\url{https://ai.meta.com/blog/llama-3-2-connect-2024-vision-edge-mobile-devices/}},
	year = {{2024}},
	note = {[Accessed 29-04-2025]},
}

@article{kazemi2024remi,
  title={Remi: A dataset for reasoning with multiple images},
  author={Kazemi, Mehran and Dikkala, Nishanth and Anand, Ankit and Devic, Petar and Dasgupta, Ishita and Liu, Fangyu and Fatemi, Bahare and Awasthi, Pranjal and Gollapudi, Sreenivas and Guo, Dee and Others},
  journal={Advances in Neural Information Processing Systems},
  volume={37},
  pages={60088--60109},
  year={2024}
}

@article{meng2024mmiu,
  title={Mmiu: Multimodal multi-image understanding for evaluating large vision-language models},
  author={Meng, Fanqing and Wang, Jin and Li, Chuanhao and Lu, Quanfeng and Tian, Hao and Liao, Jiaqi and Zhu, Xizhou and Dai, Jifeng and Qiao, Yu and Luo, Ping and Others},
  journal={arXiv preprint arXiv:2408.02718},
  year={2024}
}

@article{zhao2024benchmarking,
  title={Benchmarking multi-image understanding in vision and language models: Perception, knowledge, reasoning, and multi-hop reasoning},
  author={Zhao, Bingchen and Zong, Yongshuo and Zhang, Letian and Hospedales, Timothy},
  journal={arXiv preprint arXiv:2406.12742},
  year={2024}
}

@inproceedings{wu2024towards,
  title={Towards open-ended visual quality comparison},
  author={Wu, Haoning and Zhu, Hanwei and Zhang, Zicheng and Zhang, Erli and Chen, Chaofeng and Liao, Liang and Li, Chunyi and Wang, Annan and Sun, Wenxiu and Yan, Qiong and Others},
  booktitle={European Conference on Computer Vision},
  pages={360--377},
  year={2024},
  organization={Springer}
}

@article{liu2024convbench,
  title={ConvBench: A Multi-Turn Conversation Evaluation Benchmark with Hierarchical Ablation Capability for Large Vision-Language Models},
  author={Liu, Shuo and Ying, Kaining and Zhang, Hao and Lin, Yuqi and Zhang, Tianle and Li, Chuanhao and Qiao, Yu and Luo, Ping and Shao, Wenqi and Zhang, Kaipeng and Others},
  journal={Advances in Neural Information Processing Systems},
  volume={37},
  pages={100734--100782},
  year={2024}
}

@article{wei2022chain,
  title={Chain-of-thought prompting elicits reasoning in large language models},
  author={Wei, Jason and Wang, Xuezhi and Schuurmans, Dale and Bosma, Maarten and Xia, Fei and Chi, Ed and Le, Quoc V and Zhou, Denny and Others},
  journal={Advances in neural information processing systems},
  volume={35},
  pages={24824--24837},
  year={2022}
}

@article{li2025benchmark,
  title={Benchmark evaluations, applications, and challenges of large vision language models: A survey},
  author={Li, Zongxia and Wu, Xiyang and Du, Hongyang and Nghiem, Huy and Shi, Guangyao},
  journal={arXiv preprint arXiv:2501.02189},
  volume={1},
  year={2025}
}

@article{yu2023mm,
  title={Mm-vet: Evaluating large multimodal models for integrated capabilities},
  author={Yu, Weihao and Yang, Zhengyuan and Li, Linjie and Wang, Jianfeng and Lin, Kevin and Liu, Zicheng and Wang, Xinchao and Wang, Lijuan},
  journal={arXiv preprint arXiv:2308.02490},
  year={2023}
}

@inproceedings{masry2022chartqa,
  title={ChartQA: A Benchmark for Question Answering about Charts with Visual and Logical Reasoning},
  author={Masry, Ahmed and Do, Xuan Long and Tan, Jia Qing and Joty, Shafiq and Hoque, Enamul},
  booktitle={Findings of the Association for Computational Linguistics: ACL 2022},
  pages={2263--2279},
  year={2022}
}

@inproceedings{zitkovich2023rt,
  title={Rt-2: Vision-language-action models transfer web knowledge to robotic control},
  author={Zitkovich, Brianna and Yu, Tianhe and Xu, Sichun and Xu, Peng and Xiao, Ted and Xia, Fei and Wu, Jialin and Wohlhart, Paul and Welker, Stefan and Wahid, Ayzaan and Others},
  booktitle={Conference on Robot Learning},
  pages={2165--2183},
  year={2023},
  organization={PMLR}
}

@inproceedings{li2024manipllm,
  title={Manipllm: Embodied multimodal large language model for object-centric robotic manipulation},
  author={Li, Xiaoqi and Zhang, Mingxu and Geng, Yiran and Geng, Haoran and Long, Yuxing and Shen, Yan and Zhang, Renrui and Liu, Jiaming and Dong, Hao},
  booktitle={Proceedings of the IEEE/CVF Conference on Computer Vision and Pattern Recognition},
  pages={18061--18070},
  year={2024}
}

@inproceedings{li2024monkey,
  title={Monkey: Image resolution and text label are important things for large multi-modal models},
  author={Li, Zhang and Yang, Biao and Liu, Qiang and Ma, Zhiyin and Zhang, Shuo and Yang, Jingxu and Sun, Yabo and Liu, Yuliang and Bai, Xiang},
  booktitle={proceedings of the IEEE/CVF conference on computer vision and pattern recognition},
  pages={26763--26773},
  year={2024}
}

@misc{llamaLlama,
	author = {meta},
	title = {{L}lama --- llama.com},
	howpublished = {\url{https://www.llama.com/models/llama-3/}},
	year = {{2024}},
	note = {[Accessed 06-05-2025]},
}

@misc{bytedanceSeedreamTechnical,
	author = {{bytedance}},
	title = {{S}eedream 3.0 {T}echnical {R}eport - {P}ublications - {B}yte{D}ance {S}eed {T}eam --- seed.bytedance.com},
	howpublished = {\url{https://seed.bytedance.com/en/public_papers/seedream-3-0-technical-report?view_from=research}},
	year = {{2025}},
	note = {[Accessed 06-05-2025]},
}

@article{xue2024longvila,
  title={LongVILA: Scaling Long-Context Visual Language Models for Long Videos},
  author={Xue, Fuzhao and Chen, Yukang and Li, Dacheng and Hu, Qinghao and Zhu, Ligeng and Li, Xiuyu and Fang, Yunhao and Tang, Haotian and Yang, Shang and Liu, Zhijian and Others},
  journal={CoRR},
  year={2024}
}

@article{ge2024v2pe,
  title={V2PE: Improving Multimodal Long-Context Capability of Vision-Language Models with Variable Visual Position Encoding},
  author={Ge, Junqi and Chen, Ziyi and Lin, Jintao and Zhu, Jinguo and Liu, Xihui and Dai, Jifeng and Zhu, Xizhou},
  journal={arXiv preprint arXiv:2412.09616},
  year={2024}
}

@article{cheng2024videollama,
  title={VideoLLaMA 2: Advancing Spatial-Temporal Modeling and Audio Understanding in Video-LLMs},
  author={Cheng, Zesen and Leng, Sicong and Zhang, Hang and Xin, Yifei and Li, Xin and Chen, Guanzheng and Zhu, Yongxin and Zhang, Wenqi and Luo, Ziyang and Zhao, Deli and Others},
  journal={CoRR},
  year={2024}
}

@article{li2024llava,
  title={Llava-next-interleave: Tackling multi-image, video, and 3d in large multimodal models},
  author={Li, Feng and Zhang, Renrui and Zhang, Hao and Zhang, Yuanhan and Li, Bo and Li, Wei and Ma, Zejun and Li, Chunyuan},
  journal={arXiv preprint arXiv:2407.07895},
  year={2024}
}

@inproceedings{chen2024internvl,
  title={Internvl: Scaling up vision foundation models and aligning for generic visual-linguistic tasks},
  author={Chen, Zhe and Wu, Jiannan and Wang, Wenhai and Su, Weijie and Chen, Guo and Xing, Sen and Zhong, Muyan and Zhang, Qinglong and Zhu, Xizhou and Lu, Lewei and Others},
  booktitle={Proceedings of the IEEE/CVF conference on computer vision and pattern recognition},
  pages={24185--24198},
  year={2024}
}

@article{liu2024mia,
  title={Mia-dpo: Multi-image augmented direct preference optimization for large vision-language models},
  author={Liu, Ziyu and Zang, Yuhang and Dong, Xiaoyi and Zhang, Pan and Cao, Yuhang and Duan, Haodong and He, Conghui and Xiong, Yuanjun and Lin, Dahua and Wang, Jiaqi},
  journal={arXiv preprint arXiv:2410.17637},
  year={2024}
}

@article{lu2024deepseek,
  title={Deepseek-vl: towards real-world vision-language understanding},
  author={Lu, Haoyu and Liu, Wen and Zhang, Bo and Wang, Bingxuan and Dong, Kai and Liu, Bo and Sun, Jingxiang and Ren, Tongzheng and Li, Zhuoshu and Yang, Hao and Others},
  journal={arXiv preprint arXiv:2403.05525},
  year={2024}
}

@article{wang2024muirbench,
  title={Muirbench: A comprehensive benchmark for robust multi-image understanding},
  author={Wang, Fei and Fu, Xingyu and Huang, James Y and Li, Zekun and Liu, Qin and Liu, Xiaogeng and Ma, Mingyu Derek and Xu, Nan and Zhou, Wenxuan and Zhang, Kai and Others},
  journal={arXiv preprint arXiv:2406.09411},
  year={2024}
}

@article{dong2024internlm,
  title={Internlm-xcomposer2: Mastering free-form text-image composition and comprehension in vision-language large model},
  author={Dong, Xiaoyi and Zhang, Pan and Zang, Yuhang and Cao, Yuhang and Wang, Bin and Ouyang, Linke and Wei, Xilin and Zhang, Songyang and Duan, Haodong and Cao, Maosong and Others},
  journal={arXiv preprint arXiv:2401.16420},
  year={2024}
}

@article{jiang2024maven,
  title={Maven: An effective multi-granularity hybrid visual encoding framework for multimodal large language model},
  author={Jiang, Chaoya and Jia, Hongrui and Xu, Haiyang and Ye, Wei and Dong, Mengfan and Yan, Ming and Zhang, Ji and Huang, Fei and Zhang, Shikun},
  journal={Advances in Neural Information Processing Systems},
  volume={37},
  pages={101992--102010},
  year={2024}
}

@article{dong2024internlm4k,
  title={Internlm-xcomposer2-4khd: A pioneering large vision-language model handling resolutions from 336 pixels to 4k hd},
  author={Dong, Xiaoyi and Zhang, Pan and Zang, Yuhang and Cao, Yuhang and Wang, Bin and Ouyang, Linke and Zhang, Songyang and Duan, Haodong and Zhang, Wenwei and Li, Yining and Others},
  journal={Advances in Neural Information Processing Systems},
  volume={37},
  pages={42566--42592},
  year={2024}
}

@article{ye2024mplug,
  title={mplug-owl3: Towards long image-sequence understanding in multi-modal large language models},
  author={Ye, Jiabo and Xu, Haiyang and Liu, Haowei and Hu, Anwen and Yan, Ming and Qian, Qi and Zhang, Ji and Huang, Fei and Zhou, Jingren},
  journal={arXiv preprint arXiv:2408.04840},
  year={2024}
}

@article{wu2023q,
  title={Q-bench: A benchmark for general-purpose foundation models on low-level vision},
  author={Wu, Haoning and Zhang, Zicheng and Zhang, Erli and Chen, Chaofeng and Liao, Liang and Wang, Annan and Li, Chunyi and Sun, Wenxiu and Yan, Qiong and Zhai, Guangtao and Others},
  journal={arXiv preprint arXiv:2309.14181},
  year={2023}
}

@inproceedings{singh2019towards,
  title={Towards vqa models that can read},
  author={Singh, Amanpreet and Natarajan, Vivek and Shah, Meet and Jiang, Yu and Chen, Xinlei and Batra, Dhruv and Parikh, Devi and Rohrbach, Marcus},
  booktitle={Proceedings of the IEEE/CVF conference on computer vision and pattern recognition},
  pages={8317--8326},
  year={2019}
}

@inproceedings{marino2019ok,
  title={Ok-vqa: A visual question answering benchmark requiring external knowledge},
  author={Marino, Kenneth and Rastegari, Mohammad and Farhadi, Ali and Mottaghi, Roozbeh},
  booktitle={Proceedings of the IEEE/cvf conference on computer vision and pattern recognition},
  pages={3195--3204},
  year={2019}
}

@inproceedings{goyal2017making,
  title={Making the v in vqa matter: Elevating the role of image understanding in visual question answering},
  author={Goyal, Yash and Khot, Tejas and Summers-Stay, Douglas and Batra, Dhruv and Parikh, Devi},
  booktitle={Proceedings of the IEEE conference on computer vision and pattern recognition},
  pages={6904--6913},
  year={2017}
}

@article{wang2022internvideo,
  title={Internvideo: General video foundation models via generative and discriminative learning},
  author={Wang, Yi and Li, Kunchang and Li, Yizhuo and He, Yinan and Huang, Bingkun and Zhao, Zhiyu and Zhang, Hongjie and Xu, Jilan and Liu, Yi and Wang, Zun and Others},
  journal={arXiv preprint arXiv:2212.03191},
  year={2022}
}

@inproceedings{wang2024internvideo2,
  title={Internvideo2: Scaling foundation models for multimodal video understanding},
  author={Wang, Yi and Li, Kunchang and Li, Xinhao and Yu, Jiashuo and He, Yinan and Chen, Guo and Pei, Baoqi and Zheng, Rongkun and Wang, Zun and Shi, Yansong and Others},
  booktitle={European Conference on Computer Vision},
  pages={396--416},
  year={2024},
  organization={Springer}
}

@article{koh2024visualwebarena,
  title={Visualwebarena: Evaluating multimodal agents on realistic visual web tasks},
  author={Koh, Jing Yu and Lo, Robert and Jang, Lawrence and Duvvur, Vikram and Lim, Ming Chong and Huang, Po-Yu and Neubig, Graham and Zhou, Shuyan and Salakhutdinov, Ruslan and Fried, Daniel},
  journal={arXiv preprint arXiv:2401.13649},
  year={2024}
}

@article{tan2024devbench,
  title={Devbench: A multimodal developmental benchmark for language learning},
  author={Tan, Alvin and Yu, Chunhua and Long, Bria and Ma, Wanjing and Murray, Tonya and Silverman, Rebecca and Yeatman, Jason and Frank, Michael C},
  journal={Advances in Neural Information Processing Systems},
  volume={37},
  pages={77445--77467},
  year={2024}
}

@article{jansen2024discoveryworld,
  title={DISCOVERYWORLD: A virtual environment for developing and evaluating automated scientific discovery agents},
  author={Jansen, Peter and C{\^o}t{\'e}, Marc-Alexandre and Khot, Tushar and Bransom, Erin and Dalvi Mishra, Bhavana and Majumder, Bodhisattwa Prasad and Tafjord, Oyvind and Clark, Peter},
  journal={Advances in Neural Information Processing Systems},
  volume={37},
  pages={10088--10116},
  year={2024}
}
\appendix

In this appendix, we offer more details for the proposed VisChainBench , along with the experiment details and discussions. Specifically, Appendix A
includes our project URL and benchmark download URL. Appendix B includes details on constructing the dataset. Appendix C includes our experiment settings details. Appendix D provides the datasheets for VisChainBench.

\section{Open-source Links}
All the benchmark data and code for benchmark construction are available for viewing and download via following Link:
\begin{itemize}
\item https://huggingface.co/datasets/eyehole/VisChainBench

\label{webpage}
\end{itemize}
\section{More Details of VisChainBench}
We present the constructing details of our benchmark VisChainBench,the pipeline of our data cluser,and the prompt design for task generation and image filtering, the human annotation framework.

\subsection{Keyword Generation}
In this section, we will introduce how to prompt the LLM to generate the keyword used in task generation fig.\ref{fig:keyword}. First we manually constructed JSON task examples and then prompted them with the task description. The Keyword generation model we have tested includes Llama3.3-70B,GPT4o and Deepseek-R1.
\begin{figure}
    \centering
    \begin{tcolorbox}[colback=blue!5!white, colframe=blue!50!black,title=\textbf{Prompt For JSON file Generation},title=\textbf{Prompt For Keyword Generation}]
    I want to construct a dataset about multistep scenario reasoning. For example, for the keyword "tea," there is 1. Wash the cup and little 2 .Boil water in a kettle 3. Place tea leaves or a tea bag into the teapot or cup. This is a unique way to do this without other possible solutions. Then I will find images for these steps,So the task should also be suitable for image searching. Now I want you to generate a list of keywords for seeds to generate these tasks. For example: Car tyre(Car flat tyre replacement steps). Please generate some task keywords that meet these requirements in daily tasks.
    Just output the keywords, no need for reasoning output. I need 100 words.
\end{tcolorbox}
    \caption{Prompt For Keyword Generation}
    \label{fig:keyword}
\end{figure}
\subsection{Prompt for task description file generation}

In this section we showed the prompt used for json generation fig. \ref{fig:jsonprompt}\ref{fig:json}\ref{fig:neg}.These prompts guide the creation of structured JSON workflows that simulate real-world procedural tasks requiring ordered image selection, such as "preparing tea" or "filling tooth cavities". Each task begins with a keyword seed (e.g., "tea") and undergoes rigorous filtering to reject abstract concepts (e.g., "happiness") or scenarios lacking clear visual representations. The system enforces three core principles: (1) Multi-step necessity – tasks must involve sequential dependencies where intermediate steps affect subsequent choices; (2) Distractor design – each step includes plausible but incorrect images (e.g., showing a coffee maker during tea preparation); (3) Visual feasibility – images must depict unambiguous, web-searchable actions without overlapping interpretations. Negative examples demonstrate common pitfalls, such as ambiguous correct answers (e.g., both "hanging shirt" and "folded pants" being reasonable for clothing storage) or over-specific image descriptions that hinder retrieval.
\begin{figure}
    \centering
    \begin{tcolorbox}[colback=blue!5!white, 
colframe=blue!50!black ,title=\textbf{Prompt For JSON file Generation} ]
     
    I want you to think about a list of things that can not be finished in one step,
I am interested in tasks that require a series of steps to complete, 
where the order of steps is necessary, and where the process cannot be finished in one single action. or
I want to turn this into a multi-turn image-choosing game, adding some distracting images in each step. I will give you a keyword as a seed, You should generate the content based on the seed.
You can decide task step length by yourself, since some tasks maybe only have 2 steps, while some hard tasks require more than 5 steps.
If you feel the keyword is not suitable for task generation, you can output REJECT. However you can associate the keyword like: (The keyword health is hard to generate task,but health associate-> dentists -> task: How to fill holes in your tooth.)
You need to do things extra:
1. filter the task that is not easily to find images to describe steps, output REJECT.
2. Add distraction choices to each step.
3. The image should be easy to find on the internet. Do not contain too many elements in one photo.
4. The task can be stopped halfway finished, then you can output the images for choice. 

You should ONLY output your data in JSON format.Do not generate ``` because your output will be directly sent into json.load() function. Nothing else should be generated, except REJECT, I will show you an example:
\end{tcolorbox}
    \caption{JSON file generation prompt}
    \label{fig:jsonprompt}
\end{figure}
\begin{figure}
    \centering
    \begin{tcolorbox}[colback=blue!5!white, colframe=yellow!50!black, title=\textbf{JSON Example}]
\texttt{Based on the keyword tea:} \\
\texttt{\{} \\
\quad \texttt{"initial\_scene\_description": "You are going to make a cup of tea",} \\
\quad \texttt{"Q1": \{} \\
\quad\quad \texttt{"question": "How do you prepare the water for making tea?",} \\
\quad\quad \texttt{"task\_description": "Boil water in a kettle to use for your tea.",} \\
\quad\quad \texttt{"choices": [} \\
\quad\quad\quad \texttt{\{"image": "boiling\_water\_kettle.jpg", "correct": true\},} \\
\quad\quad\quad \texttt{\{"image": "teapot\_with\_flowers.jpg", "correct": false\},} \\
\quad\quad\quad \texttt{\{"image": "coffee\_maker.jpg", "correct": false\},} \\
\quad\quad\quad \texttt{\{"image": "plastic\_cup.jpg", "correct": false\}} \\
\quad\quad \texttt{]} \\
\quad \texttt{\},} \\
\quad \texttt{"Q2": \{} \\
\quad\quad \texttt{"question": "How should you add tea to your cup or teapot?",} \\
\quad\quad \texttt{"task\_description": "Place tea leaves or a tea bag into the teapot or cup.",} \\
\quad\quad \texttt{"choices": [} \\
\quad\quad\quad \texttt{\{"image": "tea\_bag\_in\_cup.jpg", "correct": true\},} \\
\quad\quad\quad \texttt{\{"image": "spoon\_with\_sugar.jpg", "correct": false\},} \\
\quad\quad\quad \texttt{\{"image": "instant\_coffee\_pack.jpg", "correct": false\},} \\
\quad\quad\quad \texttt{\{"image": "bottle\_of\_water.jpg", "correct": false\}} \\
\quad\quad \texttt{]} \\
\quad \texttt{\}} \\
\texttt{\}} \\[1ex]
\end{tcolorbox}
    \caption{A JSON example}
    \label{fig:json}
\end{figure}
\begin{figure}
    \centering
    \begin{tcolorbox}[colback=blue!5!white, colframe=yellow!50!black, title=\textbf{Negtive Example}]
\textbf{Bad question example:} \\
\texttt{"question": "How should you cook for meal first?",} \\
\texttt{"task\_description": "Pour milk into the glass cup.",} \\
\texttt{"choices": [} \\
\quad \texttt{\{"image": "milk\_with\_glasses.jpg", "correct": true\},} \\
\quad \texttt{\{"image": "Bread\_with\_breadmachine.jpg", "correct": false\},} \\
\quad \texttt{\{"image": "instant\_coffee\_pack.jpg", "correct": false\},} \\
\quad \texttt{\{"image": "A\_pancake\_need\_to\_heat.jpg", "correct": false\}} \\
\texttt{]} \\
\textit{(There is no absolute way to cook meal and A\_pancake\_need\_to\_heat is too abstract to express in photo, so the question is bad.)} \\[1ex]

\textbf{Bad keyword example:} \texttt{``Happiness'', ``mindfulness''} (abstract concepts, hard to visualize) \\[1ex]

\textbf{Bad task example:} \texttt{Task: How to assemble a toy car.} (parts online are often mismatched, hard to depict as a coherent scenario) \\[1ex]

\textbf{Bad image choosing example:} \\
\texttt{Step: Hang the ironed clothes.} \\
\texttt{image1: shirt on a hanger. image2: pants folded and placed on a shelf. image3: person holding a basketball.} \\
(\texttt{image1} and \texttt{image2} are both reasonable, so choice is ambiguous.) \\[1ex]

\textbf{Bad image choosing example:} \\
\texttt{Step: Connect the wires to the new light fixture.} \\
\texttt{image1: person connecting wires to a new light fixture. image2: New light fixture installed and turned on. Image 3: person holding a tube cutter.} \\
(\texttt{image2} is a possible next step after \texttt{image1}, so \texttt{image1} is not uniquely correct.) \\[1ex]

\textbf{Bad image description example:} \\
\texttt{"A person holding the new hard disk drive and preparing to install it"} \\
(description too long and specific, hard to find an accurate matching image)
\end{tcolorbox}
    \caption{Negative example in prompt}
    \label{fig:neg}
\end{figure}

\subsection{Prompt for image filtering}

\begin{tcolorbox}[colback=blue!5!white, colframe=blue!50!black,title=\textbf{Prompt For JSON file Generation},title=\textbf{Prompt for image filtering}]
You are an image choosing Agent, helping to choose the best image fitting the image descriptions. You will be given multiple images with number labels on it. Now I want you based on the text descriptions, return the most relevant image.\\
Pay attention to the distractions in the image, such as unrelated text or visual elements that don't contribute to the context of the image.
If no image matches the requirement, output: NONE \\
Example 1:\\
description: A photo of a dog.\\
\texttt{<image0>,<image1>,<image2>,<image3>}\\
Output: 2\\
\\
Example 2:\\
description: A photo of a coffee bean. \\
\texttt{<image0>}(a picture of coffee bean on tea leaves),\texttt{<image1>}(a picture of coffee bean,tea and coco),\texttt{<image2>}(a picture of coffee bean only)\\
Output: 2(Since they are all coffee beans,We want other things appear LESS in photo.)\\
\\
Example 3:\texttt{<image0>,<image1>,<image2>,<image3>}\\
Text Descriptions: a dog riding a motorcycle.\\
Output: NONE\\
\\
Now, based on the descriptions, generate which image to choose. Just the number. No other output allowed.

\end{tcolorbox}
In this section we tested Qwen2-VL-72B and Qwen2-VL-7B. The images form the search engine will be input together with the prompt. If model output NONE, the framework will return more search results for possible image choices.
\subsection{Web UI for human annotators}
\begin{figure}[h]  
    \centering
    \includegraphics[height=0.4\textheight]{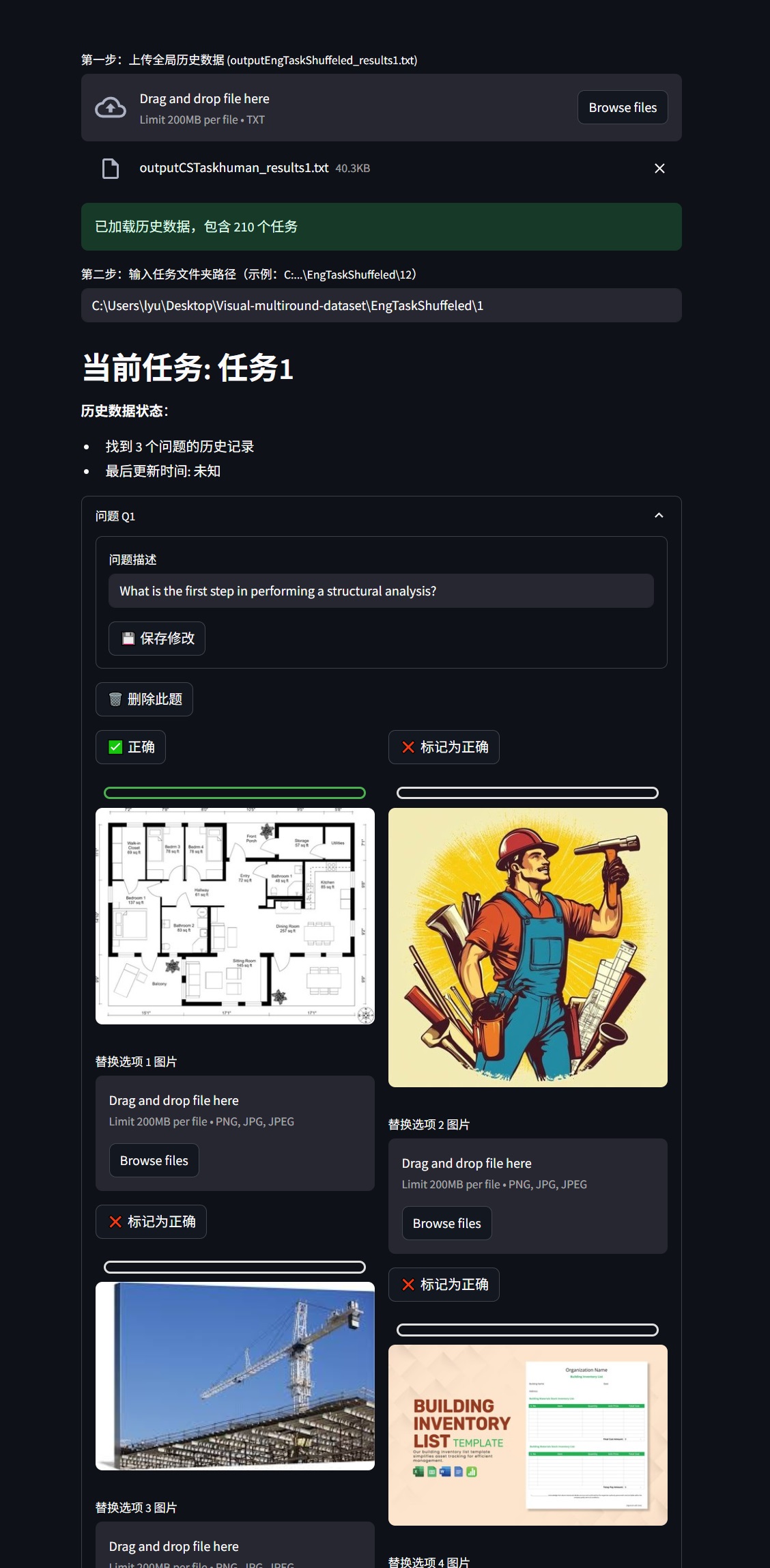}
    \caption{Human annotators webpage}
    \label{fig:web}
\end{figure}

In this section we use the streamlit framework to form a Web UI for manual data inspection and annotation.The interface is shown as Fig. \ref{fig:web}.
\section{Experiment setting details}
We initialise each interaction with a foundational prompt that explicitly defines the multimodal task. And the prompt enforces strict output formatting through structural requirements: \texttt{ANSWER: [Your choice here]}.We simply use a Regex function to extract the answer following the instructed label "ANSWER", prohibiting markdown or free-form text generation. The answer without following the instructed format will be seen as wrong. Here we only show the prompt for Image-only Multi-turn reasoning in Fig \ref{fig:test} due to the space limitations.Full prompt can be seen in our code.
\begin{figure}
\begin{tcolorbox}[colback=blue!5!white, 
colframe=blue!50!black ,title=\textbf{Prompt For Image-only Multi-turn reasoning testing} ]
You are a assistant in pure image condition task. You will be shown an initial image and a series of images representing situations and options.
For each step, you will see a condition image showing the current situation and multiple option images labeled 1, 2, 3.
Your task is to choose the most appropriate option (1, 2, or 3) for the given condition.
Your answer should begin with 'ANSWER:'.
\end{tcolorbox}
    \caption{Prompt for Image-only Multi-turn reasoning testing}
    \label{fig:test}
\end{figure}

\end{document}